\documentclass[conference]{IEEEtran}
\IEEEoverridecommandlockouts

\usepackage{amsmath,amssymb,amsfonts}
\usepackage{cite}
\usepackage{graphicx}
\usepackage{textcomp}
\usepackage{xcolor}
\usepackage{bbm}

\usepackage{enumitem}
\usepackage{tabularx}

\usepackage{times}
\usepackage{multicol}

\usepackage{graphicx}
\usepackage{bm}
\usepackage[nolist]{acronym}

\usepackage[linesnumbered,ruled]{algorithm2e}

\usepackage{mathtools}
\usepackage{array}
\usepackage{subcaption}
\usepackage{diagbox}
\usepackage{booktabs}
\usepackage{adjustbox}
\usepackage{multirow}
\usepackage{etoolbox}\AtBeginEnvironment{algorithmic}{\small}

\usepackage{soul}
\newif\ifshowcomments
 
 \showcommentstrue  

\ifshowcomments 
    \newcommand{\bae}[1]{\hl{[SB: #1]}\protect\color{black}} 
    \newcommand{\di}[1]{\hl{[DI: #1]}\protect\color{black}} 
    \newcommand{\ft}[1]{\hl{[FT: #1]}\protect\color{black}} 
    \newcommand{\jd}[1]{\hl{[JD: #1]}\protect\color{black}} 
    
\else
    \newcommand{\bae}[1]{}
    \newcommand{\di}[1]{}
    \newcommand{\ft}[1]{}
    \newcommand{\jd}[1]{}
\fi

\usepackage{lipsum}
\usepackage[colorinlistoftodos]{todonotes}

\usepackage{amsthm}

\usepackage{bbm} 
\usepackage[colorlinks = True, linkcolor = blue, citecolor = blue,hidelinks]{hyperref}

\usepackage{tikz}
\usetikzlibrary{arrows.meta,positioning,fit,calc,shapes.misc}

\usepackage[normalem]{ulem}

\title{\LARGE \bf Adaptive Time Step Flow Matching \\for Autonomous Driving Motion Planning}

\author{Ananya Trivedi$^{*1,2}$ 
\quad{Anjian Li}$^{1,3}$  \quad{Mohamed Elnoor}$^{1,4}$ \quad {Yusuf Umut Ciftci}$^{1,5}$ \quad{Avinash Singh}$^1$ \\ \quad Jovin D'sa$^1$  \quad Sangjae Bae$^1$ \quad David Isele$^1$ \quad Ta\c{s}k{\i}n Pad{\i}r$^2$ \quad Faizan M. Tariq$^{*1}$ 
\thanks{All work was done at HRI.}
\thanks{$^{1}$Honda Research Institute, San Jose, CA 95134, USA}
\thanks{$^{2}$Northeastern University, Boston, MA 02115, USA}
\thanks{$^{3}$Princeton University, Princeton, NJ 08544, USA}
\thanks{$^{4}$University of Maryland, College Park, MD 20742, USA}
\thanks{$^{5}$Stanford University, Stanford, CA 94305, USA}
\thanks{$^{*}$Corr
esponding authors: \href{mailto:trivedi.ana@northeastern.edu}{\texttt{trivedi.ana@northeastern.edu}} \& \href{mailto:faizan_tariq@honda-ri.com}{\texttt{faizan\_tariq@honda-ri.com}
}}
}

\begin{document}

\maketitle

\begin{abstract}
Autonomous driving requires reasoning about interactions with surrounding traffic. A prevailing approach is large-scale imitation learning on expert driving datasets, aimed at generalizing across diverse real-world scenarios. For online trajectory generation, such methods must operate at real-time rates. Diffusion models require hundreds of denoising steps at inference, resulting in high latency. Consistency models mitigate this issue but rely on carefully tuned noise schedules to capture the multimodal action distributions common in autonomous driving. Adapting the schedule, typically requires expensive retraining. To address these limitations, we propose a framework based on conditional flow matching that jointly predicts future motions of surrounding agents and plans the ego trajectory in real time. We train a lightweight variance estimator that selects the number of inference steps online, removing the need for retraining to balance runtime and imitation learning performance. To further enhance ride quality, we introduce a trajectory post-processing step cast as a convex quadratic program, with negligible computational overhead. Trained on the Waymo Open Motion Dataset, the framework performs maneuvers such as lane changes, cruise control, and navigating unprotected left turns without requiring scenario-specific tuning. Our method maintains a 20 Hz update rate on an NVIDIA RTX 3070 GPU, making it suitable for online deployment. Compared to transformer, diffusion, and consistency model baselines, we achieve improved trajectory smoothness and better adherence to dynamic constraints. Experiment videos and code implementations can be found at \href{https://flow-matching-self-driving.github.io/}{https://flow-matching-self-driving.github.io/}
\end{abstract}

\section{Introduction}
Autonomous driving in urban environments requires reasoning over a wide range of factors. For instance, navigating a crowded intersection may require anticipating whether other vehicles will yield for the ego to proceed. In such cases, the desired behavior can often be expressed as a set of interpretable rules that can be directly encoded into an online trajectory optimization framework~\cite{rules_based_motion-planning}. However, as the range of driving scenarios grows, the rule set can become increasingly unwieldy and hard to generalize. 

Several large-scale, open-sourced datasets such as nuScenes~\cite{nuscenes} and Waymo Open Motion~\cite{womd} capture diverse urban driving scenarios through rich sensor data, high-definition maps, and detailed agent trajectories. To mitigate the scalability limitations of rule-based planners, recent approaches have adopted imitation learning based on these datasets~\cite{goalflow}. This entails training a deep neural network capable of handling a wide range of motion planning tasks.

Diffusion models have shown strong performance in robotic manipulation, where long-horizon planning is prioritized over fast reaction times~\cite{diffusion_tri}. It has also been effectively applied to controllable traffic simulation~\cite{zhong2022guided} and open-loop robotic trajectory optimization~\cite{li2024diffusolve}. However, autonomous driving requires planners that operate at high frequencies to enable reactive decision-making. The need for potentially hundreds of denoising steps makes diffusion models impractical for real-time use~\cite{diffusion_ad}. 



Conditional flow matching has gained significant traction in the generative modeling community, with applications ranging from image generation to molecular design~\cite{basic_flow_matching,fancy_flow_matching}. It involves training a neural network to predict a continuous velocity field that transforms a known base distribution, often a Gaussian, into the data distribution under a given conditional input. During inference, this velocity field is integrated forward in time to recover the predicted output. Recent applications to motion forecasting and planning have shown that even a small number of inference steps can produce high-quality trajectories~\cite{goalflow,cfm_planninng_and_prediction}. Additionally, the number of inference steps does not need to be fixed during training. These properties make flow matching a strong and flexible candidate for motion planning tasks that demand both computational efficiency and behavioral diversity.

In this paper, we propose a flow matching based motion planning framework that predicts the future motion of surrounding agents and plans the ego vehicle’s trajectory in response. Recent work by Hu et al.~\cite{adaflow} showed that variance in the training loss of flow-matching models correlates with integration error during inference. We leverage this observation and train an auxiliary feedforward network to estimate this variance. The predicted variance determines the integration step size during online inference, using finer steps in interactive urban scenarios where the model is less confident. This removes the need for manually tuning inference schedules.

\begin{figure*}[t]
    \centering
    \includegraphics[width=0.78\textwidth]{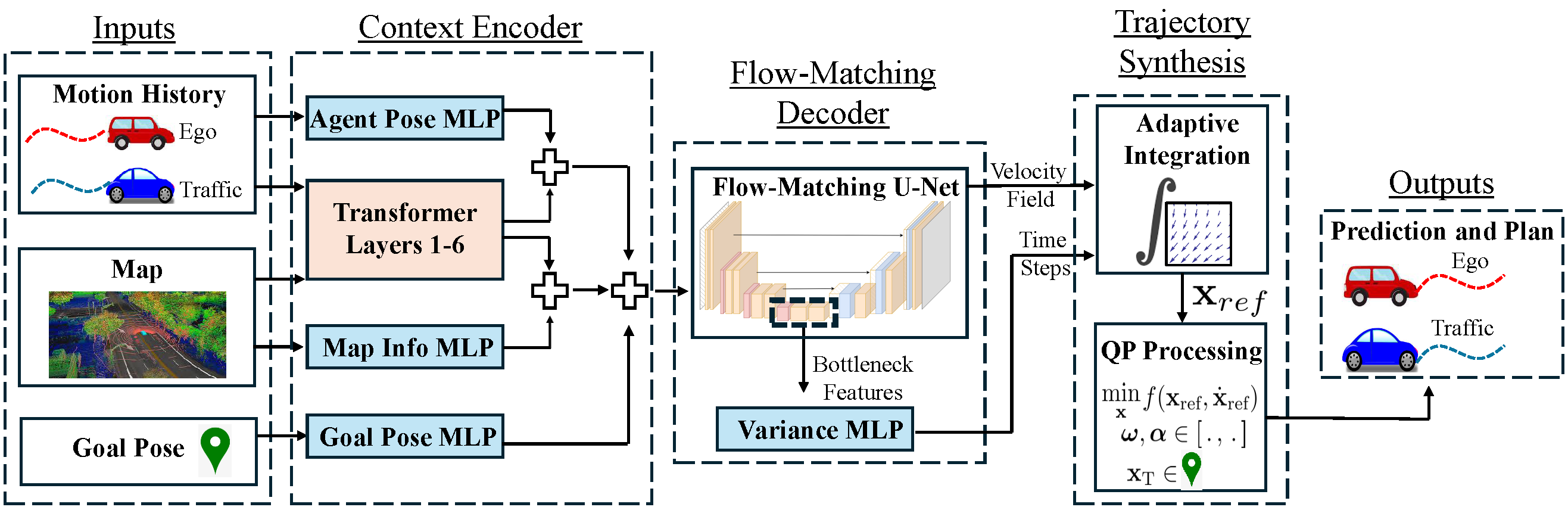}
    \caption{Our approach encodes past motion, map layout, and the desired pose using a transformer-based encoder. The resulting representation is passed to a flow matching network with an adaptive number of integration steps, followed by a lightweight post-processing step to generate a motion plan for the ego vehicle and future behavior predictions for surrounding agents.}
    \label{fig:block_diagram}
\end{figure*}

Since predicted trajectories may not perfectly match the expert demonstrations they are replicating, even small errors can often lead to motion that is dynamically inconsistent or uncomfortable for passengers~\cite{safeflowmatcher}. To address this, we introduce a post-processing step posed as a convex quadratic program (QP) that improves both ride comfort and dynamic feasibility by smoothing the trajectory and enforcing physical constraints. This additional step runs in under one millisecond, thereby introducing negligible computational overhead. An overview of the proposed method is summarized in Fig.~\ref{fig:block_diagram}.

We evaluate our approach on the Waymo Open Motion Dataset and compare it against diffusion model, consistency model, and transformer model baselines. Our method consistently outperforms these approaches in terms of trajectory smoothness, goal-reaching performance, and generalization across diverse maneuvers. We further show how, by optimizing our implementation for the inference stage via TensorRT and ONNX deployment~\cite{tensorrt}, the complete pipeline can run at approximately 20 Hz, supporting real-time deployment. To summarize, our key contributions are:
\begin{itemize}
    \item A flow-matching-based motion planner for interactive autonomous driving that handles diverse maneuvers without scenario-specific tuning and runs in real time.
    \item A variance-guided inference mechanism that adaptively selects the number of neural network function  evaluations, removing the need for tuning or retraining.
    \item A lightweight post-processing module formulated as a convex QP that enhances ride comfort and dynamic feasibility with negligible overhead.
\end{itemize}

\section{Related Works}
Traditional model-based planners formulate trajectory generation as a constrained optimization problem that incorporates vehicle dynamics and safety constraints~\cite{ad_constraints}. These methods typically rely on hand-crafted cost functions, which are difficult to tune and fail to generalize across diverse real-world scenarios~\cite{cc,modeling_dynamics}. Moreover, incorporating predictions of surrounding agents as additional costs or constraints significantly increases the computational complexity of the underlying optimization problem~\cite{trajectron}.

Recent work has explored generative models for imitating expert driving behavior directly from data. Variational Autoencoders (VAEs) generate motion plans by sampling from a learned latent space~\cite{vae}, while Generative Adversarial Networks (GANs)~\cite{gans} use a generator–discriminator setup to mimic expert behavior. Both these methods have been applied to motion planning and prediction tasks~\cite{goal_gan,vae_ad}. However, they often struggle to represent rare maneuvers present in expert demonstrations. Both VAEs and GANs can produce unrealistic outputs, often biased toward more frequent demonstrations and lacking coverage of less common but valid driving behaviors~\cite{vae_bad}. Transformer-based models~\cite{mtr} offer strong imitation learning performance but require large neural network architectures that may be too expensive for online replanning.



Diffusion models~\cite{diffusion_tri} have shown promise in planning and prediction but typically require hundreds of denoising steps, making them impractical for real-time use. To reduce this overhead, recent variants such as MotionDiffuser~\cite{motion_diffuser} use fewer denoising steps by introducing trajectory priors built from clustered examples in the training set.
This speeds up inference, but rare or long-tail behaviors are often missed, as they are typically absent from the clustered priors.

To address the inference-time inefficiency of diffusion models, prior work used consistency models~\cite{consistency_models} to develop an interactive motion planner for autonomous driving~\cite{consistency_model_hri}. These models are trained with a fixed noise schedule and enforce consistency across denoising steps during inference. The method in ~\cite{consistency_model_hri} was applied to the Waymo Open Motion Dataset, which contains diverse urban driving scenarios, and demonstrated strong performance using only four denoising steps at inference. To handle this scenario diversity, tuning the noise schedule was necessary to achieve an effective trade-off between inference speed and imitation learning accuracy. Each adjustment required retraining, thereby increasing model development time.

Trajectory-Conditional Flow Matching (T-CFM)~\cite{cfm_planninng_and_prediction} applies flow matching to aircraft trajectory forecasting and maze navigation tasks. It achieves state-of-the-art results with sampling speeds up to $100{\times}$ faster than diffusion and often requires only a single integration step. GoalFlow~\cite{goalflow} extends this idea to autonomous driving by conditioning the vector field on a goal point along with LiDAR and camera inputs. However, it is limited to single-agent settings and requires a separate goal-generation module during training.

In this paper, we use variance-adaptive flow matching to adjust the integration steps based on predicted uncertainty. This increases the number of function evaluations in regions where the model is less confident, enabling finer-grained updates in parts of the scene context that were underrepresented during training. It mitigates the bias of output trajectories toward scenes more frequently observed in the imitation learning dataset, as seen in VAEs and GANs, and avoids the reliance on fixed anchor trajectories in diffusion-based methods. This variable time-stepping scheme also eliminates the need for manually tuned noise schedules associated with consistency-based methods. Finally, by optimizing our implementation for runtime inference and incorporating a lightweight post-processing step, we achieve real-time performance unlike several transformer-based architectures with high computational overhead.

\section{Preliminaries}
\subsection{Problem Statement}
Let \(t_{\text{curr}}\) denote the current time step. With a sampling interval of \(\Delta t = 0.1\,\text{s}\), we use a past horizon of \(K = 10\) steps, to collect state sequences for all agents. Each agent’s state at time \(t\) is \(x_t \in \mathbb{R}^n\), where \(n = 4\) includes position and velocity components. The ego vehicle’s past trajectory is denoted \(S^\text{past}_e := (x^e_0 = x^e_{t_{\text{curr}}}, \ldots, x^e_{K - 1})\). Past trajectories for other agents are denoted \(S^\text{past}_\text{obj} := \{S^\text{past}_j\}_{j=1}^{N_o}\), where each \(S^\text{past}_j := (x^j_0 = x^j_{t_\text{curr}}, \ldots, x^j_{K - 1})\). We include up to the closest \(N_o = 5\) agents whose positions at \(t_{\text{curr}}\) lie within a fixed distance threshold \(R_o = 10\,\text{m}\) from the ego. The map is represented as a tensor $\mathcal{M} \in \mathbb{R}^{N_m \times L \times D}$, where $N_m$ is the number of polylines, $L$ is the number of points per polyline, and $D$ is the number of attributes per point. Each point on a polyline includes several attributes such as its position, orientation, a semantic tag indicating whether it lies on a lane centerline, road edge, crosswalk boundary, and so on. Finally, let \(s^g_e\) denote the desired ego pose at the end of the horizon. Then, given the inputs \((S^{\text{past}}_e, S^{\text{past}}_{\text{obj}}, \mathcal{M}, s^g_e)\) and a future planning horizon of \(T = 80\) timesteps, the goal is to simultaneously generate a motion plan for the ego \(S^{\text{plan}}_e := (x^e_1, \ldots, x^e_{T})\) and a prediction of the behavior of surrounding agents denoted as \(S^\text{pred}_{obj} := \{S^\text{pred}_j\}_{j=1}^{N_o}\), where each \(S^\text{pred}_{j}:= (x^j_1, \ldots, x^j_{T})\).

\subsection{Motion Transformer Encoder}
Shi et al.~\cite{mtr} proposed a transformer-based architecture for autonomous driving which used an encoder to fuse agent histories and polyline map features into scene context, and a decoder to generate multiple plausible future trajectories. In our work, we retain their encoder and use its output, denoted by \(\mathcal{C}(S^{\text{past}}_e, S^{\text{past}}_{\text{obj}}, \mathcal{M},s^g_e)\), as the conditional input to our flow matching based trajectory generator. It is worth noting that our method does not depend on a specific encoder design and can flexibly incorporate any scene encoder that provides rich contextual representations.

\subsection{Variance-Adaptive Flow Matching}
\subsubsection*{Overview of Flow Matching}
Flow matching is a generative modeling technique that transforms a sample \(z_0 \sim p_0\), typically drawn from a standard Gaussian prior, into a corresponding sample \(z_1 \sim p_1\) from the data distribution, conditioned on a context vector \(c\) that encodes relevant input information~\cite{basic_flow_matching}. This transformation is realized by training a neural network to represent a time-dependent velocity field \(v_\theta\), such that \(\frac{d}{dt} z_t = v_\theta(z_t, t \mid c)\) for \(t \in [0, 1]\). The model is trained on tuples \((c, z_0, z_1)\) extracted from the Waymo Open Motion Dataset, where \(c = \mathcal{C}(S^{\text{past}}_e, S^{\text{past}}_{\text{obj}}, \mathcal{M}, s^g_e)\) is the context embedding produced by the Motion Transformer Encoder. At inference time, we sample \(z_0\) from a standard Gaussian prior and compute the context embedding \(c\), then integrate the learned velocity field \(v_\theta\) to obtain \(S^{\text{plan}}_e\) and  \(S^\text{pred}_{obj}\)  which jointly represents the the planned trajectory for the ego vehicle and the predicted behavior of surrounding agents.

\subsubsection*{Variance-Adaptive Time-Stepping}

A specific variant of flow matching, known as variance-adaptive flow matching~\cite{adaflow}, trains a lightweight feedforward neural network \(\sigma_\phi\), alongside the velocity field \(v_\theta\). The output of this network, \(\sigma_\phi(z_t,t\mid c)\), estimates the local uncertainty of the flow based on how confidently the model has learned the velocity field for a given context, \( c = \mathcal{C}(S^{\text{past}}_e, S^{\text{past}}_{\text{obj}}, \mathcal{M},s^g_e)\). In contexts where the training data was dense, the predicted variance tends to be low, indicating that the learned velocity field is reliable. Conversely, in less frequently observed contexts, the predicted variance is high, signaling greater uncertainty about the flow. At inference time, we set the time step $\Delta t \propto 1/\sigma_{\phi}(z_t, t \mid c)$. This determines the number of evaluations of $v_{\theta}$ (NFE) needed to integrate from $t = 0$ to $1$.

\section{Methodology}
In this section, we detail the algorithmic components of our proposed planner. We begin by describing the neural network architectures used to encode the scene context and subsequently generate trajectories. We then present the training procedure used to jointly optimize these components. Finally, we introduce a convex quadratic program that post-processes the generated trajectories to improve ride comfort and enforce dynamic feasibility.

\subsection{Neural Network Architectures}
\subsubsection*{Scene Encoder Architecture}
The scene encoder processes the ego and surrounding agent histories, map polylines, and goal pose using a combination of Multi Layer Perceptrons (MLPs) and a transformer. Agent histories and map inputs are first embedded using an MLP, then passed through the transformer to capture spatial and temporal interactions. The goal pose is embedded separately and combined with the transformer outputs. The resulting token embeddings are flattened, masked, and projected into a fixed-length context vector \(c = \mathcal{C}(S^{\text{past}}_e, S^{\text{past}}_{\text{obj}}, \mathcal{M}, s^g_e)\), which conditions the flow-matching trajectory generator.


\vspace{0.5em}
\subsubsection*{Velocity Field and Variance Head Architecture}
The velocity field neural network \( v_\theta \) is implemented as U-Net with base width 128 and dimensional multipliers \((1, 2, 4)\), following the design in~\cite{unet}. It takes the conditional context embedding \( c \) as input and predicts per-timestep trajectories in \((x, y, v_x, v_y)\) space. The variance estimator \( \sigma_\phi \), described in the previous section, operates on the bottleneck features of the U-Net. These features are passed through a four-layer MLP with hidden dimension 512 and Sigmoidal Linear Unit (SiLU) activations. The network outputs a scalar variance estimate, which is used to modulate the integration step size \( \epsilon_t \) as:
\begin{equation}\label{step_size}
    \epsilon_t = \max \left( \frac{\eta}{\sigma_\phi},\; \epsilon_{\text{min}} \right),
\end{equation}
where \( \eta=0.1 \) is a user-defined regulation constant, and \( \epsilon_{\text{min}}=0.01 \) is a configurable minimum allowable step size.

\subsection{Training Procedure}
We jointly train the scene encoder, the velocity field network, and the variance prediction network. For the scene encoder, we adopt the encoder training loss \( \mathcal{L}_{\text{encoder}} \) defined in the Motion Transformer framework~\cite{mtr}.

During training, we sample $z_0 \sim p_0$, which is a standard multivariate Gaussian distribution $\mathcal{N}(0, I)$ and $z_1 \sim p_1$ is sampled from the data distribution of the normalized ground-truth future trajectory in the Waymo Open Motion Dataset (WOMD). At a uniformly sampled time \( t \in [0, 1] \), we construct the interpolated state \( z_t = (1 - t) z_0 + t z_1 \), and define the target velocity as \( z_1 - z_0 \). The context vector \( c = \mathcal{C}(S_e^{\text{past}}, S_{\text{obj}}^{\text{past}}, M, s_e^{\text{goal}}) \) is used to condition both the velocity field and variance networks, which are trained using the following loss~\cite{adaflow}:
\begin{align}
\mathcal{L}_{\text{flow}} &= \mathbb{E}_{t \sim \mathcal{U}(0, 1), \; z_0 \sim p_0, \; z_1 \sim p_1} \left[ \mathcal{L}(v_\theta, \sigma_\phi) \right] \label{eq:flow_loss} \\
\mathcal{L}(v_\theta, \sigma_\phi) &= 
\frac{\| z_1 - z_0 - v_\theta(z_t, t \mid c) \|^2}{2 \sigma_\phi(z_t, t \mid c)} 
+ \log \sigma_\phi(z_t, t \mid c) \nonumber
\end{align}

This loss encourages the velocity predictor to match the target vector field while allowing the predicted variance to increase in high-error regions and decrease in confident regions. The log term acts as a regularization penalty that prevents the predicted variances from becoming arbitrarily large. The total training loss is a weighted combination of the encoder and flow objectives:
\begin{equation}
\mathcal{L}_{\text{total}} = \beta_1 \mathcal{L}_{\text{encoder}} + \beta_2 \mathcal{L}_{\text{flow}}
\label{eq:total_loss}
\end{equation}

\subsubsection*{Constrained Refinement of Predicted Trajectories}

The trajectory generated by the adaptive flow-matching network may have sharp turns or fail to reach the target goal by the end of the prediction horizon. To improve trajectory smoothness and encourage goal reaching, we apply a post-processing step to the ego motion plan that enforces constraints on lateral acceleration, angular velocity, and final position.

Given discrete positions \((x_k, y_k)\), we compute the linear velocities and accelerations as:
\begin{align*}
    v_x^k &= \frac{x_{k+1} - x_k}{\Delta t}, \quad
    v_y^k = \frac{y_{k+1} - y_k}{\Delta t}, \\
    a_x^k &= \frac{x_{k+2} - 2x_{k+1} + x_k}{\Delta t^2}, \quad
    a_y^k = \frac{y_{k+2} - 2y_{k+1} + y_k}{\Delta t^2}.
\end{align*}
The lateral acceleration \( \alpha^k \) and angular velocity \( \omega^k \) at time \( k \) are computed as~\cite{modeling_dynamics}: 
\[
\alpha^k = \frac{a_x^k v_x^k + a_y^k v_y^k}{\sqrt{(v_x^k)^2 + (v_y^k)^2}}, \quad
\omega^k = \frac{v_x^k a_y^k - v_y^k a_x^k}{(v_x^k)^2 + (v_y^k)^2}.
\]
We enforce bounds of the form \( |\alpha^k| \leq \alpha_{\text{limit}} \), \( |\omega^k| \leq \omega_{\text{limit}} \), and require that the final position lies within a tolerance \( r \) of the desired goal. These constraints are non-convex due to bilinear dependencies on position terms. To avoid solving a computationally expensive optimization program, we linearize it around the flow matching output trajectory \(\mathbf{\tau}^{\text{ref}}=((x_0^{\text{ref}}, y_0^{\text{ref}}), \cdots, (x_{T-1}^{\text{ref}}, y_{T-1}^{\text{ref}}))\) to obtain affine surrogates \( \bar{\alpha}^k \) and \( \bar{\omega}^k \). We then impose absolute value bounds on the linearized quantities, relaxed using slack variables:
\begin{equation}
|\bar{\alpha}^k| \leq \alpha_{\text{limit}} + s_{\text{acc}}^k, \,
|\bar{\omega}^k| \leq \omega_{\text{limit}} + s_{\omega}^k, \,
s_{\text{acc}}^k \geq 0,\, s_{\omega}^k \geq 0.
\label{eq:accel_constraints}
\end{equation}
\begin{equation}
|x_T - x_{\text{goal}}| \leq r + s_{\text{goal}}^x, \,
|y_T - y_{\text{goal}}| \leq r + s_{\text{goal}}^y, \,
s_{\text{goal}}^{x,y} \geq 0,\,
\label{eq:goal_constraints}
\end{equation}

We solve the following convex quadratic program, where \(\mathbf{\tau}=((x_0, y_0), \cdots, (x_{T}, y_{T}))\) denotes the ego vehicle's optimized trajectory and \(s\) denotes all associated slack variables:
\[
\begin{aligned}
\min_{\mathbf{\tau}, s} \quad
& \underbrace{w_{\text{track}} \sum_{k=0}^{T-1} \left\|
\begin{bmatrix} x_k \\ y_k \end{bmatrix}
-
\begin{bmatrix} x^{\text{ref}}_k \\ y^{\text{ref}}_k \end{bmatrix}
\right\|^2}_{\text{Position tracking}} 
+ \underbrace{w_T \left\|
\begin{bmatrix} x_T \\ y_T \end{bmatrix}
-
\begin{bmatrix} x_{\text{goal}} \\ y_{\text{goal}} \end{bmatrix}
\right\|^2}_{\text{Goal-reaching}} \\
& + \underbrace{w_{\text{smooth}} \sum_{k=0}^{T-2} \left\|
\begin{bmatrix} x_{k+1} - x_k \\ y_{k+1} - y_k \end{bmatrix}
-
\begin{bmatrix} x^{\text{ref}}_{k+1} - x^{\text{ref}}_k \\ y^{\text{ref}}_{k+1} - y^{\text{ref}}_k \end{bmatrix}
\right\|^2}_{\text{Reference-centered velocity smoothing}} \\
& + \underbrace{w_{\text{acc}} \sum_k s_{\text{acc}}^k}_{\text{Lateral slack}} 
+ \underbrace{w_{\omega} \sum_k s_{\omega}^k}_{\text{Angular slack}} 
+ \underbrace{w_{\text{goal}} (s_{\text{goal}}^x + s_{\text{goal}}^y)}_{\text{Goal slack}}
\end{aligned}
\]
\[
\text{subject to the constraints in Eqs.~\ref{eq:accel_constraints} and~\ref{eq:goal_constraints}.}
\]

The weights \( w_{\text{track}}, w_T, w_{\text{smooth}}, w_{\text{acc}}, w_{\omega}, w_{\text{goal}} \) control the trade-off between tracking accuracy, smoothness, and constraint satisfaction. The output of this convex quadratic program is the final ego-vehicle motion plan. We solve it efficiently using off-the-shelf solvers such as OSQP~\cite{osqp} in around 1\,millisecond per trajectory.

\section{Experiments and Results}
This section presents experiments on the Waymo Open Motion Dataset. We begin with a description of the dataset and evaluation metrics, followed by qualitative visualizations that illustrate the planner’s performance across diverse real-world scenarios. We then benchmark our method against diffusion, consistency, and transformer baselines in terms of planning accuracy, constraint satisfaction, and runtime.

\subsection{Dataset and Evaluation Setup}
The Waymo Open Motion Dataset (WOMD)~\cite{womd} is a large-scale benchmark covering diverse urban driving scenarios. Each sample includes one second of past motion for all agents, eight seconds of future ground-truth trajectories, and high-definition maps. Our planner is conditioned on the final pose of the ego vehicle from the ground-truth, though this can be replaced by any goal generation module, such as~\cite{goalflow}. All evaluations are performed on the official validation split, which is further categorized into interactive and non-interactive subsets based on agent interaction levels.

To evaluate planner performance, we report metrics that capture both alignment with ground-truth trajectories and ride quality. These include minimum average displacement error (minADE) and final displacement error (minFDE), which measure average and final deviations from the ground truth. Goal-reaching error is defined as the Euclidean distance between the final pose of the planned trajectory and the assigned goal. To assess ride quality, we compute angle change, total path length, and average curvature, capturing directional smoothness, path efficiency, and turning sharpness, respectively. Constraint satisfaction is measured through violations in lateral acceleration, angular velocity, and goal-reaching constraints. We also report collision rate, defined as the percentage of scenarios where the ego comes within 2 meters of another agent’s ground-truth trajectory. All metrics are computed in an open-loop setting and averaged over validation subsets.

\subsection{Trajectory Output Visualizations}
\begin{figure}[t]
    \centering
    \begin{subfigure}{0.495\linewidth}
        \centering
        \includegraphics[width=0.8\linewidth]{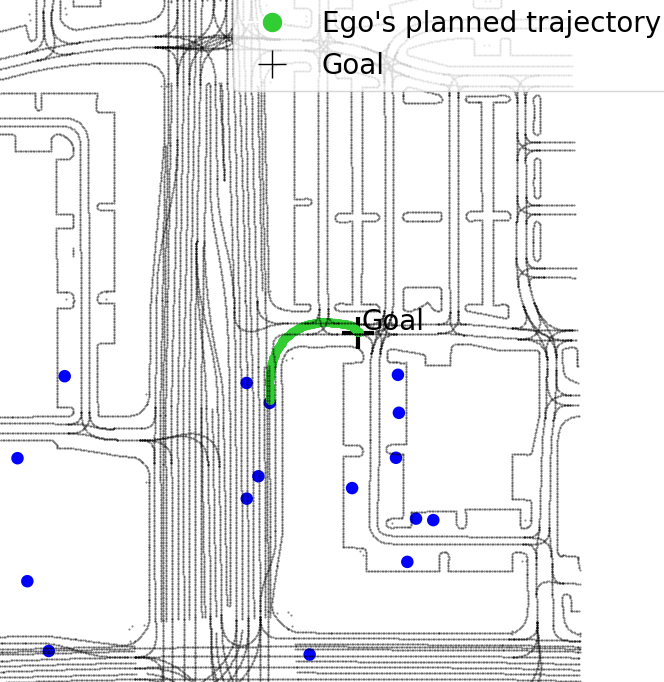}
        \caption{Original Goal}
    \end{subfigure}\hfill
    \begin{subfigure}{0.495\linewidth}
        \centering
        \includegraphics[width=0.8\linewidth]{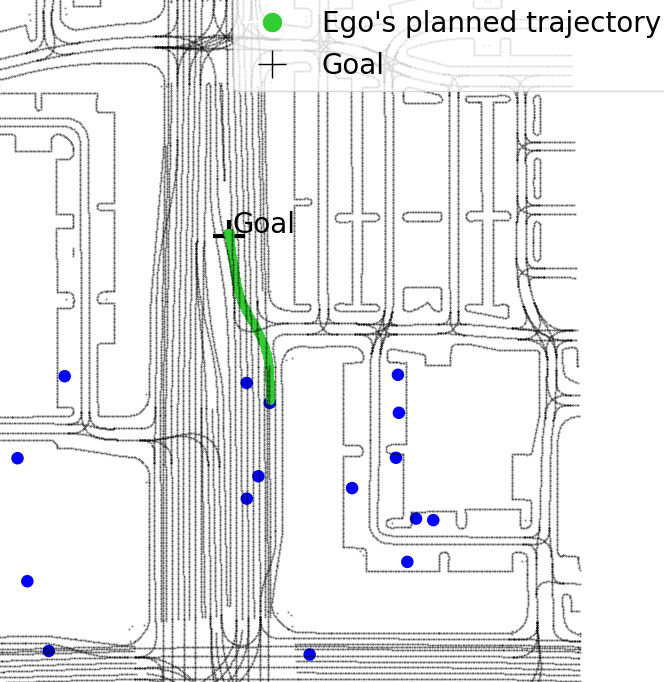}
        \caption{Updated Goal}
    \end{subfigure}
    \caption{(a) The ego takes a sharp right exit. (b) From the same initial pose, the goal is changed to a left lane change. The policy adapts and produces smooth, lane-aligned trajectories.}
    \label{fig:goal_change}
\end{figure}

\begin{figure*}[t]
    \centering
    \includegraphics[width=0.9\textwidth]{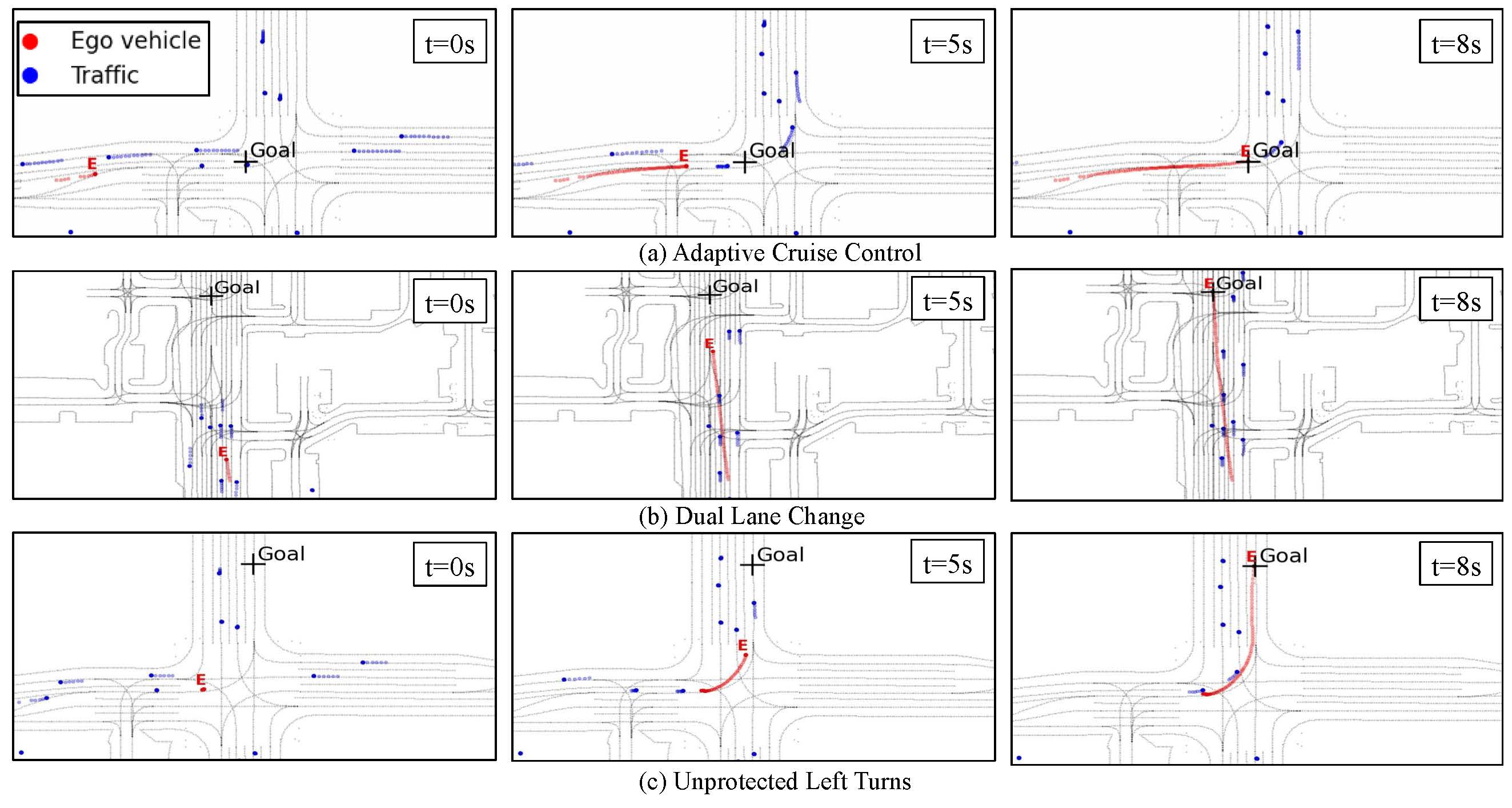}
    \vspace{0.6em}
    \includegraphics[width=0.9\textwidth]{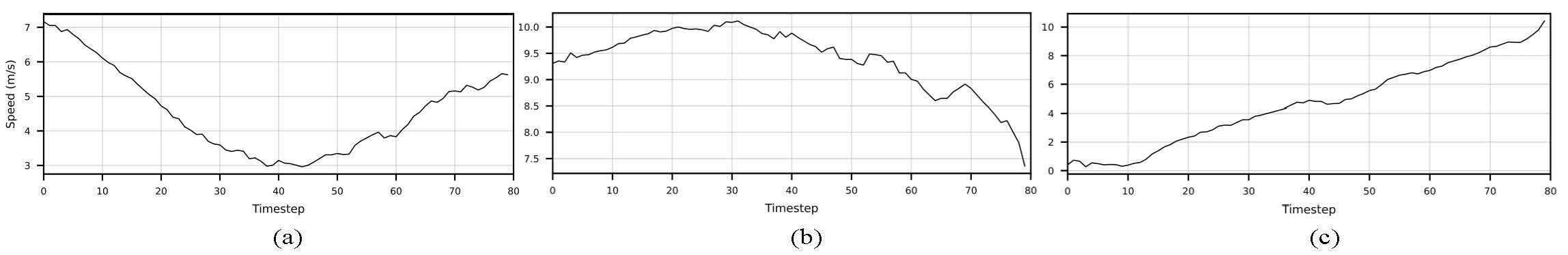}
    \caption{Example maneuvers handled by the proposed method: (a) approaching stop-and-go traffic, (b) changing multiple lanes in front of traffic to reach the goal, and (c) yielding to oncoming vehicles before executing an unprotected left turn. The bottom row shows the corresponding ego speed profiles. Videos of these and additional maneuvers can be found at \href{https://flow-matching-self-driving.github.io/}{https://flow-matching-self-driving.github.io/}}

    \label{fig:adaflow_snapshots_speed}
\end{figure*}

In a receding-horizon implementation, where the planner is invoked at regular intervals with updated scene context, it is essential that the goal-conditioned policy can reliably adapt to changes in the desired end pose. To stress test this behavior, we evaluate the planner from a fixed initial condition while specifying two distinct goals: one that requires a sharp right exit and another that demands a left lane change. As shown in Fig.~\ref{fig:goal_change}, the planner successfully adapts to both target poses, producing smooth trajectories that remain consistent with the road layout.

We showcase the planner’s  behavior in three challenging real-world scenarios: (a) adaptive cruise control, (b) dual lane change, and (c) unprotected left turn. Each row in Fig.~\ref{fig:adaflow_snapshots_speed} shows map snapshots at three time steps (0s, 5s, 8s), depicting the ego trajectory, surrounding traffic, and the goal. The ego's full trajectory history is visualized, while only 1 second of history is shown for other agents to reduce clutter. The bottom row presents the ego's speed profile for each maneuver.

In (a), the ego approaches stationary traffic, slows down accordingly, and resumes acceleration once the traffic begins to move, maintaining a safe following distance. In (b), the ego accelerates to merge ahead of traffic and performs two lane changes to reach the goal, then eases its speed near the endpoint. In (c), the ego waits for oncoming vehicles to pass, then gradually accelerates to complete an unprotected left turn. All of these behaviors emerge from the same unified framework without requiring any scenario-specific tuning. In contrast, RL- or MPC-based planners typically require separate offline training or online optimization for each maneuver~\cite{ad_constraints}.

\begin{table}[t]
\centering
\caption{Trajectory Accuracy for \textbf{Interactive} Split}
\label{tab:interactive_realism}
\begin{tabular}{lcc}
\toprule
\textbf{Method} & \textbf{minADE} $\downarrow$ & \textbf{minFDE} $\downarrow$ \\
\midrule
Ours (Flow-Adaptive) & 0.92 & \textbf{0.09} \\ 
Flow-Euler-5  & 0.93 & 0.10 \\
Flow-Euler-50 & 1.03 & 0.09 \\
Flow-Adaptive-NoQP & 1.01 & 0.40 \\
Consistency-Guided       & \textbf{0.69} & 0.33 \\
Transformer       & 2.89 & 1.09 \\
DDPM-10           & 1.04 & 2.45 \\
DDPM-20           & 1.10 & 2.21 \\
DDIM              & 5.98 & 14.48 \\
\bottomrule
\end{tabular}
\end{table}

\begin{table}[t]
\centering
\caption{Trajectory Accuracy for \textbf{Non-Interactive} Split}
\label{tab:nonint_realism}
\begin{tabular}{lcc}
\toprule
\textbf{Method} & \textbf{minADE} $\downarrow$ & \textbf{minFDE} $\downarrow$ \\
\midrule
Ours (Flow-Adaptive) & 1.61 & \textbf{0.11} \\
Flow-Euler-5  & 1.61 & 0.13 \\
Flow-Euler-50 & 1.74 & 0.12 \\
Flow-Adaptive-NoQP & 1.70 & 0.81 \\
Consistency-Guided       & \textbf{1.23} & 0.59 \\
Transformer       & 2.80 & 9.25 \\
DDPM-10           & 1.60 & 12.64 \\
DDPM-20           & 1.60 & 11.26 \\
DDIM              & 6.25 & 22.89 \\
\bottomrule
\end{tabular}
\end{table}

\begin{table*}[t]
\centering
\caption{Trajectory Quality and Constraint Violations on the \textbf{Interactive} Split}
\label{tab:quality_and_constraints_int}
\begin{adjustbox}{max width=\textwidth}
\begin{tabular}{lccccccc}
\toprule
\textbf{Method} & \textbf{Angle Change} $\downarrow$ & \textbf{Path Length} $\downarrow$ & \textbf{Curvature} $\downarrow$ & \textbf{Collision (\%)} $\downarrow$ & \textbf{Goal Error} $\downarrow$ & \textbf{Acc. Violation} $\downarrow$ & \textbf{$\omega$ Violation} $\downarrow$ \\
\midrule
Ours (Flow-Adaptive) & \textbf{0.31} & \textbf{54.93} & \textbf{0.48} & 1.1 & \textbf{0.09} & \textbf{0.12} & \textbf{0.71} \\
Flow-Euler-5  & 0.41 & 55.09 & 0.86 & 1.53 & 0.10 & 0.36 & 1.13 \\
Flow-Euler-50 & 0.35 & 55.04 & 0.71 & 1.40 & 0.09 & 0.28 & 0.96 \\
Flow-Adaptive-NoQP & 0.69 & 56.52 & 1.07 & 1.55 & 0.67 & 2.11 & 1.32  \\ 
Consistency-Guided       & 0.65 & 55.39 & 0.97 & \textbf{0.8}  & 0.33 & 3.82 & 1.47 \\
Transformer       & 0.47 & 55.88 & 0.60 & 4.8  & 8.11 & 0.44 & 1.12 \\
DDPM-10           & 0.63 & 67.74 & 1.38 & 1.8  & 2.45 & 32.61 & 2.25 \\
DDPM-20           & 0.56 & 77.91 & 1.16 & 1.6  & 2.21 & 56.69 & 2.17 \\
DDIM              & 0.67 & 134.50 & 1.71 & 18.6 & 14.48 & 167.77 & 4.51 \\
\bottomrule
\end{tabular}
\end{adjustbox}
\end{table*}

\begin{table*}[t]
\centering
\caption{Trajectory Quality and Constraint Violations on the \textbf{Non-Interactive} Split}
\label{tab:quality_and_constraints_non_int}
\begin{adjustbox}{max width=\textwidth}
\begin{tabular}{lccccccc}
\toprule
\textbf{Method} & \textbf{Angle Change} $\downarrow$ & \textbf{Path Length} $\downarrow$ & \textbf{Curvature} $\downarrow$ & \textbf{Collision (\%)} $\downarrow$ & \textbf{Goal Error} $\downarrow$ & \textbf{Acc. Violation} $\downarrow$ & \textbf{$\omega$ Violation} $\downarrow$ \\
\midrule
Ours (Flow-Adaptive) & \textbf{0.29} & \textbf{57.05} & \textbf{0.55} & 2.4 & \textbf{0.11} & \textbf{0.31} & \textbf{0.77} \\
Flow-Euler-5  & 0.38 & 57.30 & 0.93 & 2.5 & 0.13 & 0.64 & 1.20 \\
Flow-Euler-50 & 0.33 & 57.20 & 0.77 & 2.6 & 0.12 & 0.52 & 1.01 \\
Flow-Adaptive-NoQP & 0.73 & 59.31 & 1.23 & 3.1 & 1.47 & 3.57 & 1.39  \\ 
Consistency-Guided       & 0.62 & 58.07 & 1.00 & \textbf{2.0} & 0.59 & 4.75 & 1.50 \\
Transformer       & 0.42 & 59.62 & 0.56 & 3.3 & 9.26 & 0.57 & 1.02 \\
DDPM-10           & 0.66 & 88.13 & 3.45 & 3.7 & 12.64 & 78.47 & 3.34 \\
DDPM-20           & 0.59 & 93.19 & 2.98 & 3.3 & 11.26 & 89.54 & 2.96 \\
DDIM              & 0.68 & 158.94 & 1.97 & 17.6 & 22.89 & 217.28 & 5.20 \\
\bottomrule
\end{tabular}
\end{adjustbox}
\end{table*}

\subsection{Quantitative Comparison of Different Models}
We benchmark our planner against Transformer, Diffusion, and Consistency-based baselines, all using the same context encoder shown in Fig.~\ref{fig:block_diagram}. The Transformer baseline follows the MTR decoder~\cite{mtr} and does not condition on the ego goal. In contrast, the Diffusion, Consistency, and our models use goal-conditioned U-Net decoders with the same encoder. For diffusion-based models, we evaluate the DDPM baseline~\cite{vanilla_diffusion} with 10 and 20 denoising steps, as well as the deterministic DDIM variant~\cite{ddim}. The Consistency model, adapted from~\cite{consistency_model_hri}, performs four forward passes followed by a gradient guided constraint enforcement step. To ensure fair runtime comparison, we cap the number of gradient descent iterations to match the 20 Hz planning rate achieved by our method. We evaluate three variants of our approach: Flow-Euler, which uses a fixed number of integration steps (5 and 50); Flow-Adaptive, which adjusts the step size based on a learned variance predictor; and Flow-Adaptive–NoQP, which omits the QP refinement. Both Flow-Euler and Flow-Adaptive methods include the QP-based post-processing step. All U-Net decoders are exported to ONNX and accelerated using TensorRT~\cite{tensorrt}.

As shown in Tables~\ref{tab:interactive_realism} and~\ref{tab:nonint_realism}, our adaptive flow matching method achieves the lowest minFDE across both interactive and non-interactive scenarios. This improvement is due to the learned velocity field and the  enforcement of goal satisfaction during post-processing. The final displacement error of $0.11m$ is close to the goal-reaching tolerance of $r=0.1m$ used in Eq.~\ref{eq:flow_loss}, indicating effective constraint satisfaction. Our method reports a slightly higher minADE than the consistency model, reflecting an inherent tradeoff: the post-processing step may shift the trajectory away from the nominal flow matching output to better satisfy constraints. Since ADE is computed relative to the ground truth, such deviations can slightly increase intermediate errors.

Tables~\ref{tab:quality_and_constraints_int} and~\ref{tab:quality_and_constraints_non_int} evaluate trajectory smoothness and constraint satisfaction across all models. Our method outperforms all baselines on nearly all metrics in both interactive and non-interactive settings, achieving lower angular change, shorter path length, reduced curvature, smaller goal-reaching error, and fewer violations of acceleration and angular velocity limits. These gains are primarily due to the learned flow matching planner, which produces smooth, goal-directed trajectories. The quadratic program provides a lightweight refinement that further reduces constraint violations. The only exception is the collision rate, where our method performs comparably to the best baseline, the Consistency model.
\begin{figure}[t]
    \centering
    \includegraphics[width=0.7\linewidth]{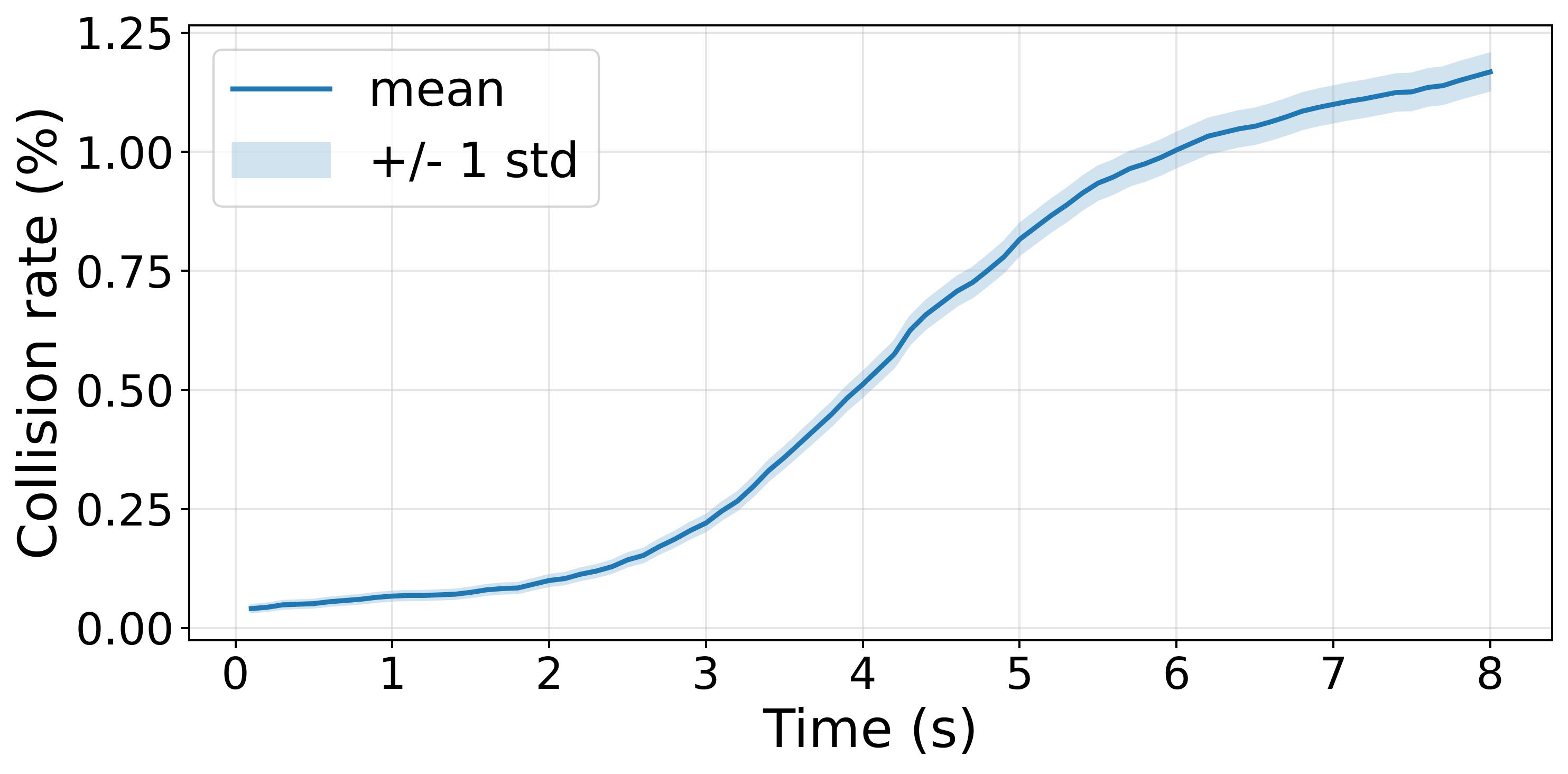}
    \caption{Collision rate plot: Collisions are rare initially and higher toward the end of the horizon. With continuous replanning, the early time steps are most critical for safety.}
    \label{fig:cum_collision_rate}
\end{figure}

Since collision rate is a key safety measure, we analyze collision occurrences along the planning horizon of $8$ seconds. Figure~\ref{fig:cum_collision_rate} shows that most collisions occur near the end rather than at the beginning. This is expected, as small deviations from the ground-truth trajectory accumulate over time, leading to larger divergence later in the plan. While collisions are measured in an open-loop setting without replanning, real-world deployment would involve continuous updates based on the evolving scene context. In such cases, the planner would regenerate the trajectory at each cycle, making safety in the early steps more critical. Our method maintains collision-free behavior in these initial segments. Finally, future work will involve incorporating obstacle avoidance as explicit constraints in the post-processing optimization, as shown in ~\cite{safeflowmatcher}.

\begin{table}[t]
\centering
\caption{Inference Time and NFE Analysis}
\label{tab:inference_nfe}
\begin{tabular}{@{}lcc@{}}
\toprule
\textbf{Method} & \textbf{Time (ms)} $\downarrow$ & \textbf{NFE} $\downarrow$ \\
\midrule
Ours (Flow-Adaptive)      & 45 ± 4.9   & 4.7 ± 1.4 \\
Flow-Euler-5              & 55 ± 3.8   & 5.0 ± 0.0 \\
Flow-Euler-50             & 676 ± 25.3 & 50.0 ± 0.0 \\
Flow-Adaptive-NoQP        & 43 ± 4.2   & 4.7 ± 1.4 \\
Consistency-Guided          & 52 ± 4.5   & 4.0 ± 0.0 \\
Transformer               & 37 ± 3.1   & N/A \\
DDPM-10                   & 83 ± 7.2   & 10.0 ± 0.0 \\
DDPM-20                   & 125 ± 10.7 & 20.0 ± 0.0 \\
DDIM                      & \textbf{37 ± 2.5} & \textbf{4.0 ± 0.0} \\
\bottomrule
\end{tabular}
\end{table}


\subsection{Analysis of Inference Time and Adaptive Integration}
\begin{figure}[t]
    \centering
    \includegraphics[width=0.6\linewidth]{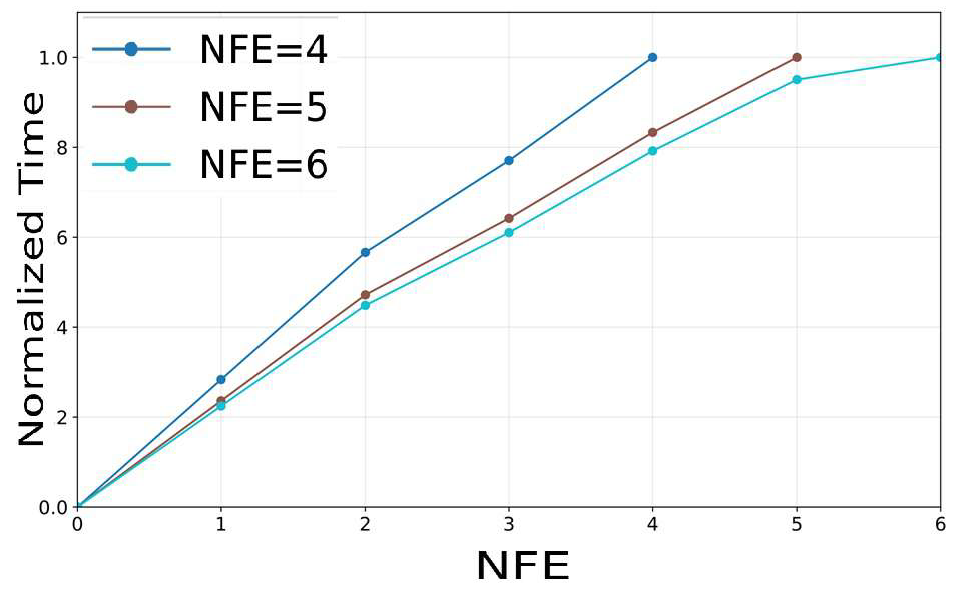}
    \caption{Normalized integration time for different NFE. The non-uniform progression reflects variance-based step size adjustment, unlike the uniform fixed-step Euler integration.}
    \label{fig:adaptive_step_size}
\end{figure}
Table~\ref{tab:inference_nfe} reports the mean and standard deviation of inference time and number of U-Net function evaluations (NFE) for all methods, computed over all scenes in the interactive validation split. Inference time is measured as the total time needed to produce traffic behavior predictions and ego plans from the motion history, map, and goal inputs.

Our method achieves an average inference time of 45 milliseconds, corresponding to a 20 Hz planning rate suitable for real-time deployment. While DDIM is slightly faster at 37 milliseconds, its performance in trajectory quality and constraint satisfaction falls short of ours. Other baselines such as Consistency, Transformer, and DDPM-10 operate in a similar time range but do not match our overall metrics.

To visualize how adaptive time stepping operates, we analyze the average integration step size during inference for NFE values of 4, 5, and 6 on the interactive validation set. As shown in Fig.~\ref{fig:adaptive_step_size}, we observe non-uniform step sizes, indicating that the variance module allocates computation based on confidence. Low predicted variance suggests that the model has encountered similar scene contexts during training and assigns larger steps due to a reliable velocity field. Conversely, high variance results in smaller steps in unfamiliar scene contexts.

To further illustrate the benefit of adaptive integration, we compare against fixed-step variants of our model. We observe that flow matching with 50 steps achieves performance comparable to the adaptive variant but at significantly higher computational cost due to increased U-Net evaluations. We also evaluate flow matching with 5 steps, selected to match the average number of function evaluations used by the adaptive model. This variant performs notably worse, indicating that fixed low-resolution integration fails to generalize across diverse scenarios. We include the Flow-Adaptive–NoQP variant, which excludes the post-processing step. This exhibits more frequent constraint violations. As shown in Table~\ref{tab:inference_nfe}, the QP refinement adds only about 1-2 milliseconds to the runtime while significantly improving motion planning performance.

Finally, it is worth noting that in~\cite{consistency_model_hri}, the Consistency model required extensive tuning of the noise schedule to optimize performance on the WOMD. This process involved retraining both the encoder and U-Net, taking approximately five days on four H100 GPUs per run. In contrast, our adaptive variant achieves a similar average NFE of 4.7 without any manual tuning, as the number of U-Net evaluations is automatically determined by the learned variance estimator.

\section{Conclusion and Future Work}

In this work, we presented a framework for interactive motion planning for autonomous vehicles in urban settings, based on adaptive flow matching. The approach operates in real time, requires no scenario-specific tuning, and generates smooth trajectories that try to satisfy ride quality constraints.

While our current implementation relies on polyline-based map inputs provided by the WOMD, this assumes access to dedicated preprocessing and sensor infrastructure. As future work, we aim to replace the map encoder with vision-based alternatives that operate directly on raw LiDAR or camera data~\cite{vision_encoder}. We also plan to evaluate the planner in closed-loop simulation environments such as MetaDrive~\cite{metadrive}, followed by deployment on real hardware platforms.

\bibliographystyle{IEEEtran}
\bibliography{refs}






\end{document}